\documentclass[sigconf]{acmart}

\def\BibTeX{{\rm B\kern-.05em{\sc i\kern-.025em b}\kern-.08emT\kern-.1667em\lower.7ex\hbox{E}\kern-.125emX}}
    
\copyrightyear{2019}
\acmYear{2019}
\setcopyright{acmlicensed}
\acmConference[KDD 2019 Workshop]{AI for fashion
The fourth international workshop on fashion and KDD}{August 2019}{Anchorage, Alaska - USA}
\acmBooktitle{KDD 2019 Workshop: AI for fashion
The fourth international workshop on fashion and KDD, August 2019, Anchorage, Alaska - USA}
\acmPrice{15.00}
\acmDOI{10.1145/1122445.1122456}
\acmISBN{978-1-4503-9999-9/18/06}

\begin{document}

%
\title{Teaching DNNs to design fast fashion}

%
\author{Abhinav Ravi}
\email{abhinav.ravi@myntra.com}
\affiliation{%
  \institution{Myntra Designs}
  \city{Bangalore, India}
}

\author{Arun Patro}
\email{arun.patro@myntra.com}
\affiliation{%
  \institution{Myntra Designs}
  \city{Bangalore, India}
  }

\author{Vikram Garg}
\email{vikram.garg@myntra.com}
\affiliation{%
  \institution{Myntra Designs}
  \city{Bangalore, India}
}

\author{Anoop Kolar Rajagopal}
\email{anoop.kr@myntra.com}
\affiliation{%
 \institution{Myntra Designs}
 \city{Bangalore, India}
 }
 
\author{Aruna Rajan}
\email{aruna.rajan@myntra.com}
\affiliation{%
  \institution{Myntra Designs}
  \city{Bangalore, India}
  }

\author{Rajdeep Hazra Banerjee}
\email{rajdeep.banerjee@myntra.com}
\affiliation{%
  \institution{Myntra Designs}
  \city{Bangalore, India}
  }



%
\renewcommand{\shortauthors}{Abhinav and Arun, et al.}

%
\begin{abstract}
``Fast Fashion'' spearheads the biggest disruption in fashion that enabled to engineer resilient supply chains to quickly respond to changing fashion trends. The conventional design process in commercial manufacturing is often fed through ``trends''  or  prevailing  modes of dressing around the world that indicate sudden interest in a new form of expression, cyclic patterns, and popular modes of expression for a given time frame. In this work, we propose a fully automated system to explore, detect, and finally synthesize trends in fashion into design elements by designing representative prototypes of apparel given time series signals generated from social media feeds. Our system is envisioned to be the first step in design of Fast Fashion where the production cycle for clothes from design inception to manufacturing is meant to be rapid and responsive to current ``trends''. It also works to reduce wastage in fashion production by taking in customer feedback on sellability at the time of design generation. We also provide an interface wherein the designers can play with multiple trending styles in fashion and visualize designs as interpolations of elements of these styles. We aim to aid the creative process through generating interesting and inspiring combinations for a designer to mull by running them through her key customers.
\end{abstract}

%
%


%

%
\begin{teaserfigure}
  \includegraphics[width=\textwidth]{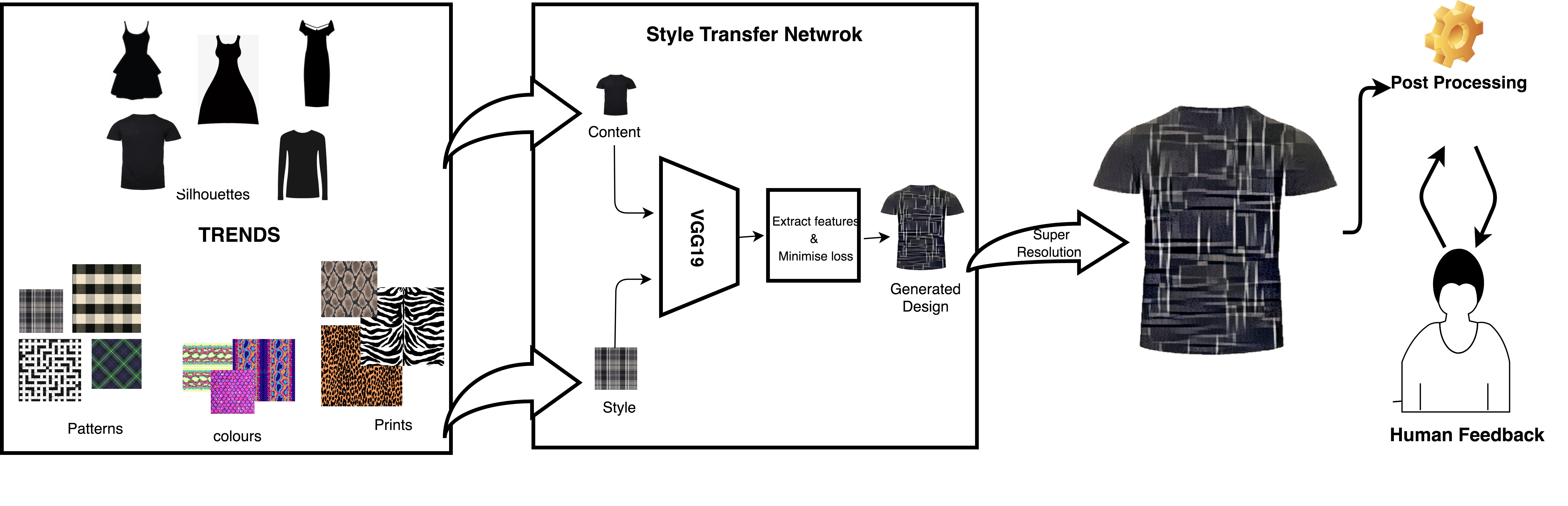}
  \caption{Schematic flow for Fast Fashion design.}
  \label{fig:teaser}
\end{teaserfigure}

%
\maketitle

\section{Introduction}
The most recent disruption in fashion manufacturing came from a movement referred to as ``Fast Fashion'' that is the ability to quickly respond to changing customer tastes and social and cultural norms of the day~\cite{fastFashion}. Fast fashion does generate a lot of waste and it has come under heavy criticism recently for its driving of fast paced consumerism, and related environmental hazards from heavy wastage. In order to  rethink fashion commercially to serve its artistic purpose of creating expressive value while staying relevant, we took a hard look at what causes wastage in the fashion supply chain, and how can we maintain diversity of fashion choice while simultaneously reducing wastage. The scope of our work is defined by what computational and programmatic advances in AI can achieve to address these challenges, over what textile engineering and better sourcing practices can achieve for culling environmental waste. While the latter methods are more holistic and address waste at the root, we aim to use systemic advances in AI to generate only relevant prototypes in fast fashion and reduce the need to produce large volumes. 

Fashion design can be inspired by runways and fashion weeks, edited fashion content offline and on the internet, street fashion, and celebrities and influencers. Staying relevant to the current, and yet allowing for diversity of expression is what a fashion designer grapples with most. Her prototypes are sketches and story boards that she then socializes with brand managers and buyers, who in turn work out the commercial implications and aim to hedge their bets by diversifying their portfolio. We noticed that the process of identifying relevance or trends that matter is highly subjective, and given the lack of support in terms of data, hedging is the most commonly used strategy, with fail-safe techniques employed to scale down complexity in design and appeal to a wider range of tastes. The entire supply chain therefore, works to hedge against originality yet to produce ``new'' content, and this becomes antithetical to the original purpose. 

We propose to reduce the number of steps and number of bets taken in this supply chain by solving for the following: 

\begin{enumerate}
    \item A data backed method to find what's relevant in the world of fashion today
    \item A quick prototyping and visualisation system that is able to extract key elements of current trends and interpolate between such elements to generate ``new'' prototypes to inspire designers.
    \item A distributed system to collect initial responses and ratings from prospective or important customers at the design stage itself, after a designer has designed to the stories supported by (1) with the help of (2)~\cite{originalityFashion}.
\end{enumerate}

Using our fast fashion prototype generation system we show that it's possible to detect the signals from the fashion trends and create the fast fashion using style transfer methods. We also collect initial responses and reviews of both designers and potential customers. 

\section{Related work}
New fashion generation is an area of recent research interest and intense activity given computational advances in deep learning.  Prutha Date et al. \cite{ganesan2017fashioning} use neural style transfer and construct approaches to create new custom clothes based on a user's preference and by learning the user's fashion choices from their current collection of clothes.
This method learns user preferences based on the user's choices or set of clothes from the user's closet. Then, it uses style transfer to generate new clothes. Hence, both styles and content (shapes) are user provided rather than being part of global trends or relevant phenomena in the world. Such a method is restrictive to a user's current sense of style, and assumes that style is not influenced by external factors. Fast fashion however rests on the premise that style is an evolving concept and is deeply influenced by current events across the world in a highly internet connected economy. Hence, we choose to focus on what's happening in the world at large rather than what a particular user may like in keeping with his or her current tastes. This also creates ``new'' and distinct elements of style rather than repetitive and banal for every user.

Some of the elements of fashion our system automatically detects from online fashion content are shapes, silhouettes, colours, prints and patterns, and textures that trend. These elements are synthesised using style transfer based techniques to generate a prototype of an original or unseen design.

One of the most popular fashion style generators published \cite{jiang2017fashion} has been optimised to preserve stylistic elements while transferring a style to content (apparel type in our case). As the authors mentioned, one of the shortcomings of their method was that plain images without content did not get modified in the applied style transformation. Starting design (computationally) as a blank slate, would therefore not work in this case. Neural style transfer by  Gatys' et al.\cite{gatys2016image} is an algorithm that combines the content representation of one image and style representation of another image to obtain a new image having the content characteristics of the first image and style characteristics of the second image. This seems like a good way to interpolate between stylistic content and geometric/structural content in fashion. This method provides the mechanism ($\alpha$ and $\beta$ weighting parameters for content loss and style loss respectively) to control the extent of content and style images which should be transferred to the final image, making it our method of choice. 
 
 Some later works \cite{sheng2018avatar}, \cite{luan2018deep} build on the work of Gatys' et al. to add features on top of vanilla style transfer techniques like multiple style integration, video stylization, localised style transfer etc.

Other work \cite{choi2014fast} \cite{beheshti2015survey} to reduce fast fashion wastage and increase choice available to customers explore the demand forecasting and prediction in the fast fashion markets. In our understanding, there is no existing published work on generating fashion prototypes that are fed by external trends. 

\section{Fast fashion generation pipeline}
Figure~\ref{fig:teaser} shows our fashion generation system. We first use google trends, our own assortment forecasting model, social media and competitive intelligence to determine the trending styles or images. From trends' signals thus detected, we find stylistic design elements such as silhouettes and pass the plain silhouette to our downstream generation pipelines as content images. We then pass the new colours or prints or patterns as style images for application of ``style transfer''~\cite{gatys2016image} to content. In order to speed up the computationally intensive style transfer step, we first apply style transfer to a low res image and then perform super resolution to scale it up for visualisation purposes before production. These high resolution images are then post processed by our designers to add their creative touches and re-imagine the final product by using our images as basis for inspiration. When their design is complete, we pass the design onto a feedback gathering system that is able to survey and sample initial responses from a wide customer base. This feedback can potentially inform the designer as to what elements of style have percolated down the fashion value chain, and what his or her customer values most. 

Our system also aids design by allowing the designer to generate several permutations and combinations of silhouettes, prints, patterns and colours for her material for inspiration. We acknowledge that creativity is more than just permuting elements of style, which is why we include designers as key owners of the fashion design process. Permuting elements of style however can serve as a source of diverse inspiration, and in combining unlikely elements of style for visualisation, a designer can be inspired to think ``out of the box''. We detail each component of our schema in further subsections.


\subsection{Trend detection} \label{ssec:num1}
We crawl Instagram feeds from various fashion influencers identified by our key customers and designers at Myntra-Jabong \cite{fashioninfluencer}. We represent images taken from these feeds as a set of visual features, by generating embeddings such as vgg-19~\cite{russakovsky2015imagenet} which is a 19 layer pretrained network trained on the ImageNet data-set\cite{simonyan2014very}. These networks and representations were chosen as they lend themselves very well to article type and further attribute type classification in fashion. We represent each Instagram feed as a time series over a set of image features, and by aggregating over several such Instagram feeds, we obtain a combined time series of world fashion in deep learned image features. 

\begin{figure*}[ht]
\centering
\includegraphics[height=1.6in, width=2\columnwidth]{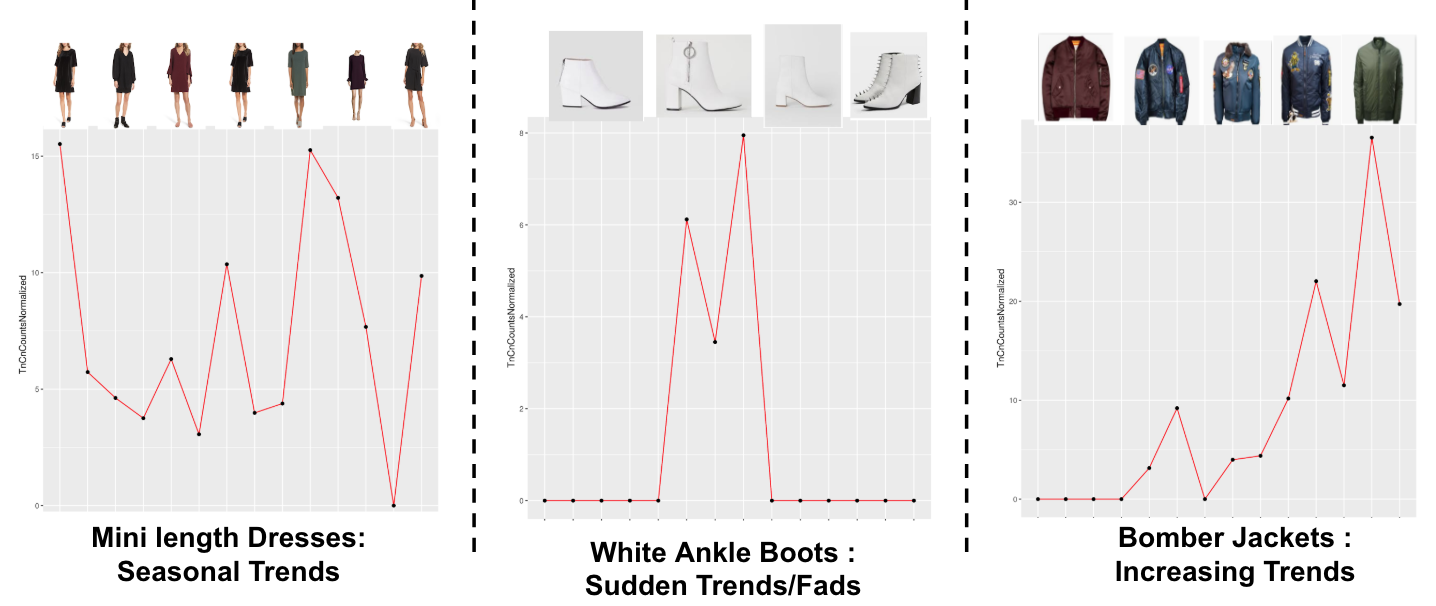}
\caption{Trend Clusters that was found by our system from Social media. Left figure shows the seasonal/core trends, middle one shows anomalous trend and right figure shows the increasing trends.}
\label{trends}
\end{figure*}

In this space we cluster the image features and look for the following characteristics 
\begin{itemize}
    \item Clusters with the largest membership - this indicates the images features which are being worn by a large number of people across the world. These are usually the core designs which are always in popular fashion magazines such as vogue, Cosmopolitan etc.
    \item Clusters of image features that show increasing growth over time - these usually indicate new design elements that come into the market.
    \item Anomalous or new clusters of image features across time - these are quirky designs usually worn by a community of people.
\end{itemize}
Figure~\ref{trends} shows representative samples and the normalized number of posts (TnCnCountNormalized) by the influencers for the cluster vs Time 
Hence, we detect trends in the raw images themselves, of new/unseen colours and patterns, popular or frequently posted colours and patterns, and features showing gradual increase in interest over time. Through this systemic detection of trends, we obtain a set of representative images and image features for trends. We further segregate the representation of these images into colours, patterns or textures, prints, and silhouettes or shapes. 

In this work, with social media data gathered over the last few years, some of the trends which we captured for the current season (Spring Summer 19) in print and pattern type are animal prints\cite{animalPrint}, camouflage print, and plaid checks. In shapes, we detected distressed jeans, boyfriend fit jeans, plaid raincoats, sports stripes etc.


\subsection{Fashion Generation}

\begin{figure*}[ht]
\centering
\includegraphics[height=2.4in]{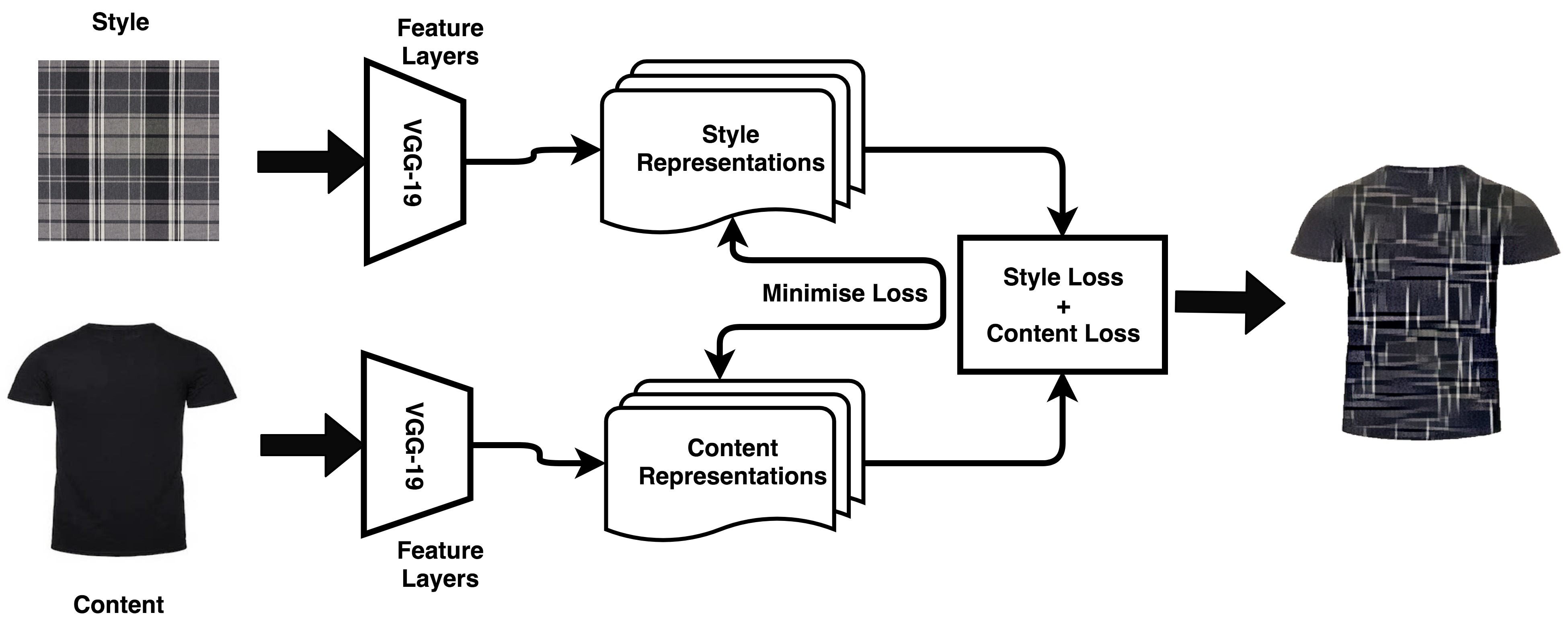}
\caption{Style transfer technique applied on silhouette and pattern.}
\label{styletransfer}
\end{figure*}

For our fashion generation pipeline, we took the vgg19 representations of identified trending images and pass the resulting representations to the style transfer algorithm ~\cite{gatys2016image}. The images of trending silhouettes were passed as content images. Trending colours, patterns or textures and prints were passed as style images. \\A gram matrix was used to represent the style transfer loss. 
The intuition behind using gram matrix is to obtain a feature map which tells us about the magnitude of correlation among the style feature representations from a particular style layer. \\
The style loss ($loss_{style}$) in equation \ref{eq:1} ~\cite{gatys2016image} signifies how different the final generated image is from the style representation of a given style image $a$.
We also measure the content loss ($loss_{content}$) which tells us how different the final generated image looks from content image $p$ (or in our case the chosen shapes or outlines for clothing) as shown in equation \ref{eq:1}. Finally, we minimize these combined losses ($loss_{total}$) and come up with the combined output image. See Figure~\ref{styletransfer}.

    A noise image $x$ is taken as the seed image and enhanced after each iteration as in the ~\cite{gatys2016image} to obtain the final generated image as per equation \ref{eq:1}.



\begin{equation} \label{eq:1}
    loss_{total} (p, a, x) = \alpha * loss_{content} (p, x) + \beta * loss_{style} (a, x)
\end{equation}

In the equation \ref{eq:1}, $x$ is the noise image, $a$ is the style image and $p$ is the content image. $\alpha$ and $\beta$ are the weighting factors which decide the contribution of these individual losses towards total combined losses ($loss_{total}$) to be minimized. $\alpha$ was maintained to be around 0.05 and $\beta$ at 5.0 for our experiments, based on observed convergence. We also used a L-BFGS optimizer that gave qualitatively better looking images (validated through designer feedback) than other faster optimizers like Adam~\cite{kingma2014adam}.

\subsection{Super Resolution}
The outputs from the style transfer algorithm are enhanced through Super Resolution using Deep Convolutional Networks (SRCNN) \cite{srcnn}. SRCNNs directly learns an end-to-end mapping between the low/high-resolution images. The model also provides a flag for noise removal that suppresses heavy gradients. The output of such images are clear, and smooth color variations.

For our experiments, we have used a scaling factor of $4$ to finally obtain a $2400x2400$ pixels image which qualifies for being around $400$ DPI. 

\section{Experiments and Results} \label{System design}
\begin{figure*}[ht]
\centering
\includegraphics[height=3.8in,width=1.8\columnwidth]{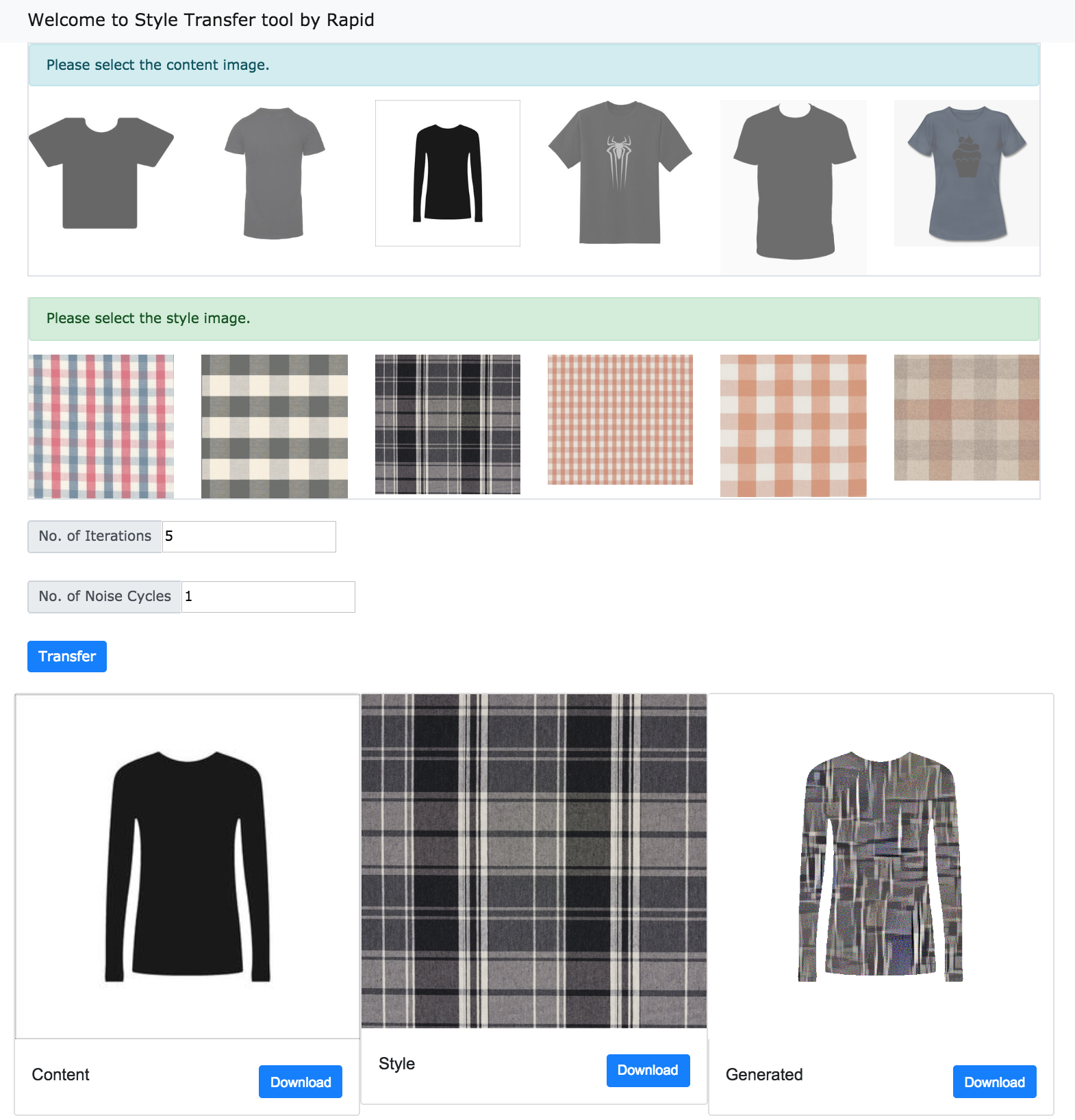}
\caption{User Interface exposed to the designers.}
\label{ui}
\end{figure*}

In our system, compute units consisted of a NVIDIA K80 GPU with 4 cores and 61GB RAM. Initially the fashion designs were generated in 600 * 600 pixels resolution. The style transfer network took about 2 minutes per image at this resolution and network configuration, to complete style transfer to content. In our system, we also allow the designer to replace the style or content image through images taken from their own Instagram feed/Pinterest board/or other images collected by them as inspirational material. The SRCNNs were highly scalable and showed sub-second latency for the scale up needed for visualisation of output. See Fig. \ref{ui} for actual system interface.

We also allow designers or users of our tool to control the iterations required for style transfer and super resolution - so that they may choose an image that represents visually interesting features to them, and can increase the number of iterations until varied representations are generated. Through experimentation with design users, we found that around 10 to 25 iterations are good enough for generating clear enough designs to be used by designers. 

Using different noise patterns helps us in generating different nuances of patterns in the generated designs as shown for a sample silhouette design with uniform noise, sinusoidal noise with one cycle and ten cycles.
The options to choose the noise cycle is provided to the designers in the web application as shown Figure~\ref{ui}, however this is more of a developer feature than a creative designer feature. 

\subsection{Generated examples}
\begin{figure*}[ht]
\centering
\includegraphics[scale=0.18]{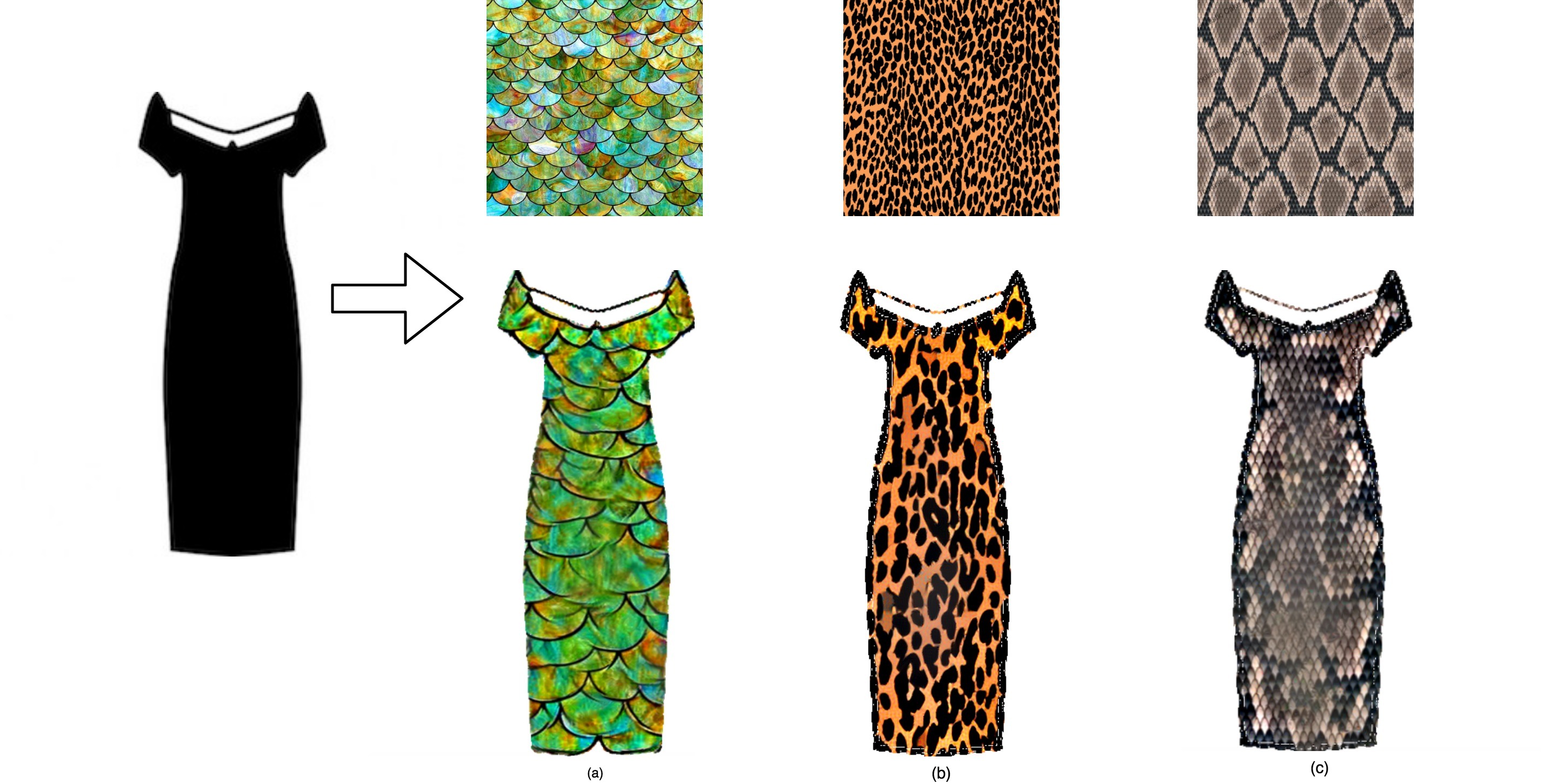}
\caption{Applying trend animal prints while keeping other parameters constant.}
\label{prints}
\end{figure*}

Figure~\ref{prints} shows results of applying different types of animal prints viz. fish scale print, leopard print and snake print to the sheath shape silhouette.

\begin{figure*}[ht]
\centering
\includegraphics[scale=0.18]{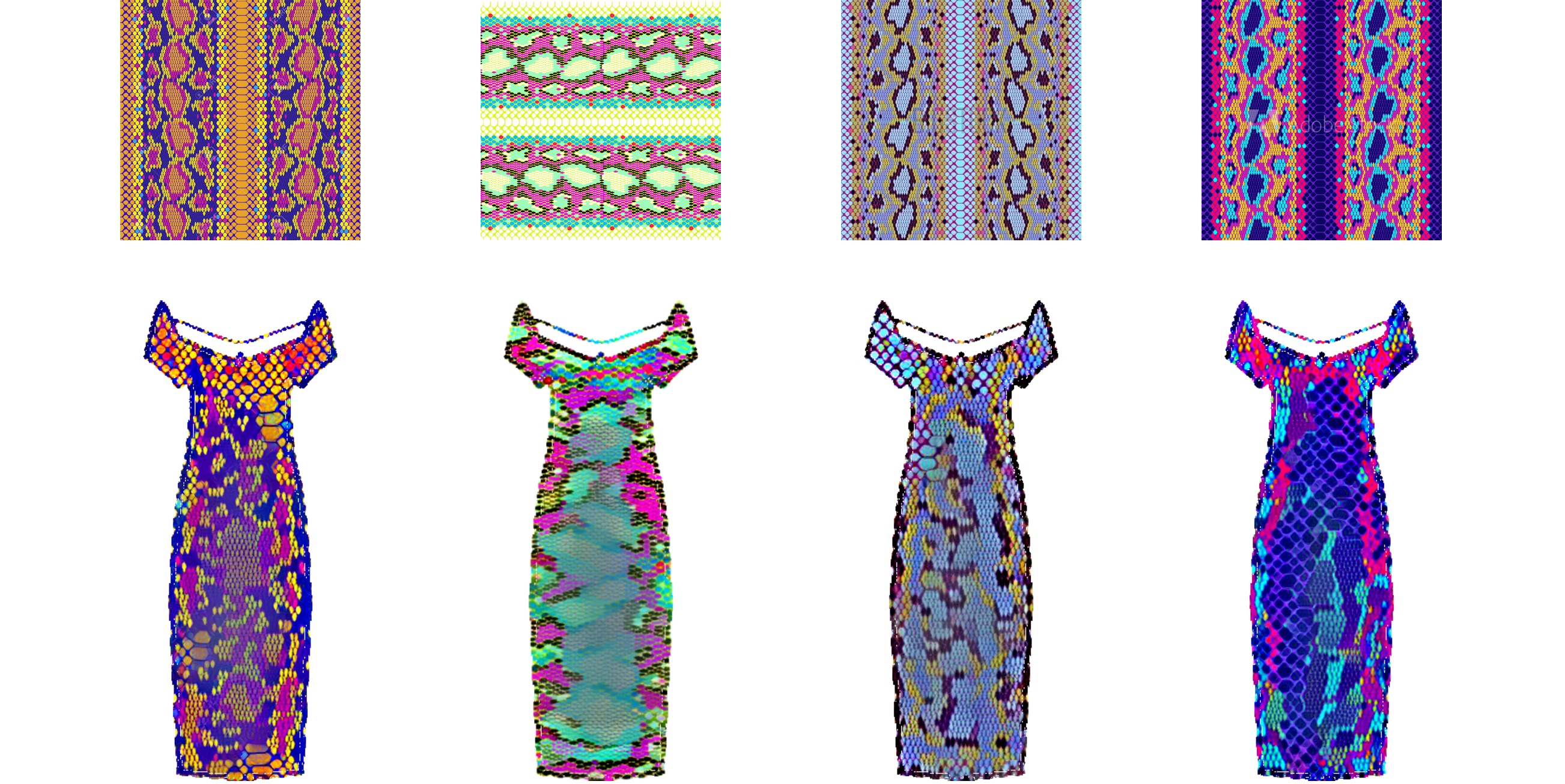}
\caption{Applying different trend colours while keeping other parameters constant.}
\label{neonColours}
\end{figure*}

Figure~\ref{neonColours} lists down the generated design while applying neon and fluorescent colours (detected as trending colours for the season) for a style image of snake print. Thus we keep same sheath shape silhouette and snake print and just change the colours of the prints.



\begin{figure*}[ht]
\centering
\includegraphics[scale=0.08]{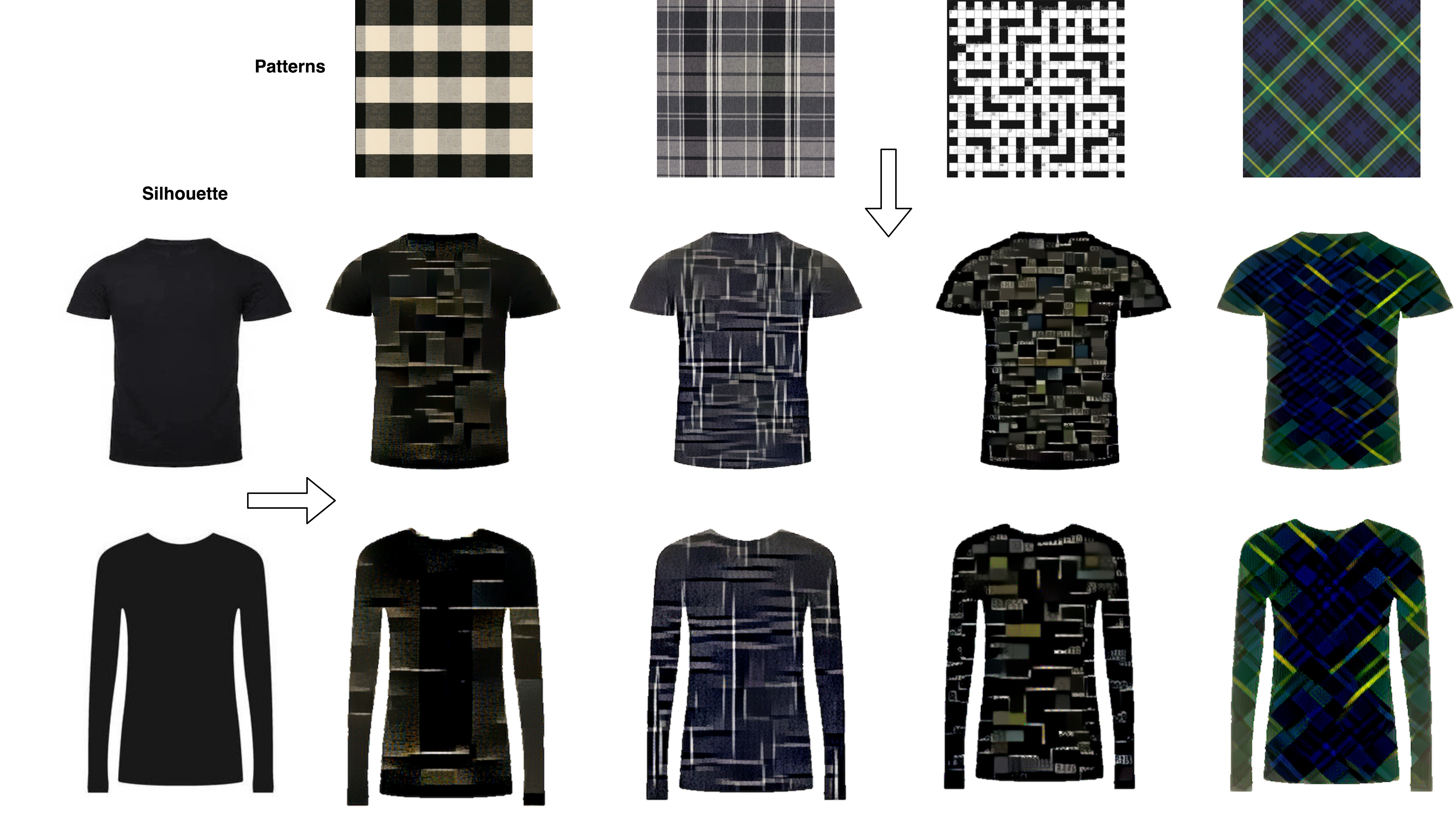}
\caption{Applying different trend patterns to different silhouettes.}
\label{tshirts}
\end{figure*}

Figure~\ref{tshirts} shows the generated designs after applying checkered and crisscross pattern to two different t-shirt silhouette.

\subsection{Evaluation}
 We conducted a questionnaire experiment to qualitatively evaluate a sample of randomly drawn 15 designs from our set of generated designs (7 women's, 8 men's) where we asked two kinds of participants to give us feedback on two major questions. For designers,(i) "Whether a particular design interests/inspires them" and (ii) "Would they process it for manufacturing to their line?" We asked end customers of fashion (picked among volunteering employees and customers of Myntra-Jabong) to rate the styles for their (i) distinctiveness or freshness in fashion content, and (ii) for whether the user would purchase them (given favourable merchandising parameters such as pricing/quality, etc). 
 
 The design participants of this experiment were experienced fashion designers with medium to high level experience (from 1 year to 6 years). 9 among these designers were male and 6 others were female participants. They were asked to rate the generated designs on a 1 to 5 scale, 1 being very poor and 5 being excellent, and also leave unstructured text as comments. End customers (10 in number) were asked to rate designs for their ability to interest/inspire, and the customer's likelihood of purchasing them, again on a scale of 1 to 5 similarly arranged as before. 
 
 \begin{figure*}[ht]
\centering
\includegraphics[scale=0.08]{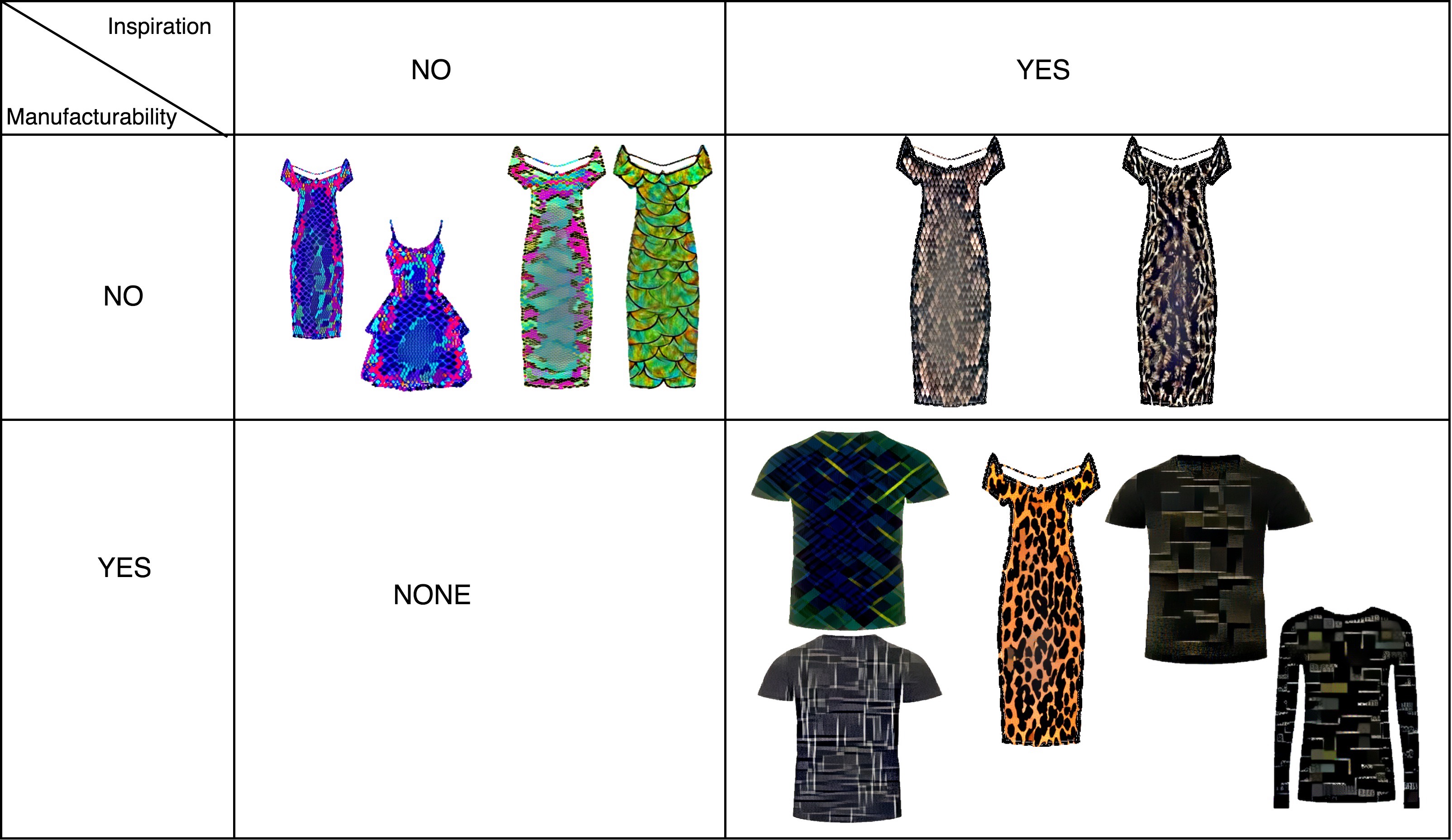}
\caption{Results segregation of the designs for questionnaire.}
\label{result}
\end{figure*}

\begin{figure*}[ht]
\centering
\includegraphics[scale=0.4]{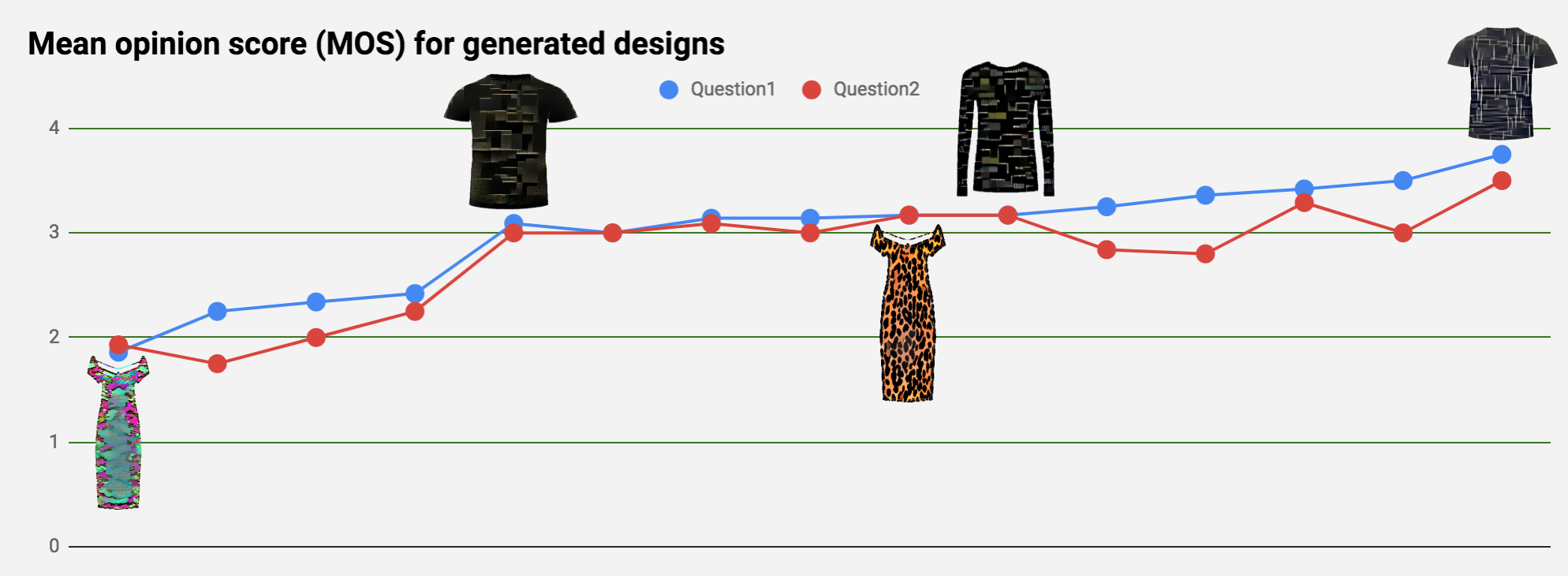}
\caption{Mean opinion score distribution of the designs from questionnaire.}
\label{mos}
\end{figure*}

For the purposes of visualisation in binaries, we regard a question as answered in affirmative if the mean score awarded was greater than $3$. 

$6$ out of these $15$ designs got affirmative score in both the questions. $4$ designs got negative score in both the questions and $5$ designs got affirmative score for first question while negative score for second question. None of the designs got negative score for inspiration components while getting an affirmative score for second question. We have shown the answers and designs in a tabular form in Figure~\ref{result}

The designs which got low inspiration and manufacturability score had one theme in common, i.e. print and colour combination. Though the snake print, neon and fluorescent bright colours were captured as part of being trendy, yet they were given the lowest score by participants. Our designers suggested that the neon and fluorescent bright colours applied on snake prints were very over powering. Interestingly, end customers rated these designs as ''fresh,'' or unseen, possibly because they deviated from all commonly accepted (sampled in our lot of $15$) schools of design thinking in fashion. The designers still maintained that neon colours with bold prints are better designed as trims over main garments for women's wear. 
Figure~\ref{prints}(a) was again called out by customers as quirky and new, while designers rejected them for the boldness of colours. 

When we scale our system across Myntra-Jabong's designer base and representative customers, we can disrupt the way fast fashion is generated today to automatically seed data backed trends, and quickly generate visual prototypes or drafts of the final apparel to be manufactured.

We also found interesting examples of creativity demonstrated by deep neural nets. In Figure~\ref{prints}(a), the semicircular fish scales adapt to sheath shaped silhouette near sleeves, neck, hemline and give it a unique mermaid look. A ruffled look at the sleeves and a narrow nape produces a mermaid like rendering, where as our network was only seeded with an image of fish scale print, and not an actual fish/aquamarine creature whose arrangement of scales it has managed to mimic. Of particular curious interest was the cut at the hemline of the sheath dress, to mimic fish scales. While the content image contained a firmly outlined sheath dress with a straight hem, the creation component of our system has interpreted the hem differently given fish scales as patterns.  
\section{Conclusion}
Through this work, we have demonstrated how the true potential of deep neural networks in disrupting businesses has only reached the tip of the iceberg. We have come out with a new system architecture for doing fast fashion which serves and augments designers creative side. We have shown several results of our individual components and have built a User interface for the designers. There are several areas ripe 
for disruption through application of deep learning methods, not just in prediction but in value creation as well. 

\bibliographystyle{ACM-Reference-Format}
\bibliography{sample-base}

%
\appendix









\end{document}